%% file: root.tex
\documentclass[letterpaper, 10 pt, conference]{ieeeconf}  

\IEEEoverridecommandlockouts                              

\usepackage{graphicx}
\usepackage{amsmath,amssymb} 
\usepackage{color}
\usepackage{mathtools}
\usepackage{commath}
\usepackage[sc,osf]{mathpazo}
\usepackage{algorithm}
\usepackage{algorithmic}
\usepackage{multirow}
\usepackage[table,xcdraw]{xcolor}
\usepackage[normalem]{ulem}
\useunder{\uline}{\ul}{}
\definecolor{red}{rgb}{0.95,0.4,0.4}
\definecolor{blue}{rgb}{0.4,0.4,0.95}
\definecolor{darkblue}{rgb}{0,0,0.8}
\definecolor{darkred}{rgb}{0.8,0,0}
\definecolor{darkgreen}{rgb}{0,0.5,0}
\definecolor{grey}{rgb}{0.6,0.6,0.6}
\definecolor{col1}{RGB}{232, 161, 148}
\definecolor{col2}{RGB}{148, 187, 232}

\title{\LARGE \bf
ADAADepth: Adapting Data Augmentation and Attention for Self-Supervised Monocular Depth Estimation
}

\author{Vinay Kaushik$^{1}$, Kartik Jindgar$^{2}$ and Brejesh Lall$^{3}$
\thanks{$^{1}$Vinay Kaushik and $^{3}$Brejesh Lall are with the Department of Electrical Engineering, IIT Delhi, India.%
$^{2}$Kartik Jindgar is with Manipal University, Jaipur.
{\tt\small \{Vinay.Kaushik,brejesh\}@ee.iitd.ac.in kjindgar@gmail.com}}%
}
\begin{document}
\maketitle
\thispagestyle{empty}
\pagestyle{empty}

\begin{abstract}
   Self-supervised learning of depth has been a highly studied topic of research as it alleviates the requirement of having ground truth annotations for predicting depth. Depth is learnt as an intermediate solution to the task of view synthesis, utilising warped photometric consistency. Although it gives good results when trained using stereo data, the predicted depth is still sensitive to noise, illumination changes and specular reflections. Also, occlusion can be tackled better by learning depth from a single camera. We propose ADAA, utilising depth augmentation as depth supervision for learning accurate and robust depth. We propose a relational self-attention module that learns rich contextual features and further enhances depth results. We also optimize the auto-masking strategy across all losses by enforcing L1 regularisation over mask. Our novel progressive training strategy first learns depth at a lower resolution and then progresses to the original resolution with slight training. We utilise a ResNet18 encoder, learning features for prediction of both depth and pose. We evaluate our predicted depth on the standard KITTI driving dataset and achieve state-of-the-art results for monocular depth estimation whilst having significantly lower number of trainable parameters in our deep learning framework. We also evaluate our model on Make3D dataset showing better generalization than other methods.
\end{abstract}

\section{Introduction}
Depth from a single image has been of utmost importance in computer vision community with the advent of deep learning. Depth prediction provides solutions for several applications including smart mobility \cite{Mauri2020DeepLF}, smartphone AR \cite{Valentin2018DepthFM}, 3D zooming \cite{Bello2019Deep3N}, face anti-spoofing \cite{Liu2019AuroraGR}, image dehazing \cite{Kim2020ImageDM}, etc.  Humans are able to perceive depth in the visible world by utilising cues like occlusion, texture differences, relative scale of neighbouring objects, lighting and shading variations along with object semantics.

Multi-view and stereo methods are computationally expensive and have high memory overheads. Depth from single image drastically reduces these complexities and is favourable for real-time systems. Deep learning provides the tools to predict depth from a single image by transforming the task into a learning problem\cite{eigen2014depth,fu2018deep}, given the ground truth depth annotations. However, capturing vast amount of ground truth data in different scenarios is a formidable task. Self-supervision for computing depth eliminates this limitation by utilising photometric warp for learning depth\cite{godard2017unsupervised,garg2016unsupervised}.

Learning from a monocular sequence is challenging due to scale ambiguity and unknown camera pose. Thus, there's an explicit need to compute camera egomotion\cite{godard2019digging,yin2018geonet}. The necessity of joint learning for depth and egomotion means that the quality of depth is highly dependent on the correctness of camera pose. Also, static scene assumption in self-supervised learning paradigm leads to holes and aberrations in pixels belonging to a moving object in the scene. Occlusions at image boundaries makes it difficult to learn depth near boundary regions (bottom image region in a forward moving camera).
Although there have been innovations in deep learning architectures\cite{guizilini20203d,johnston2020self}, loss functions\cite{godard2017unsupervised,godard2019digging}, masking strategies \cite{vijayanarasimhan2017sfm,zhou2017unsupervised,luo2018every,godard2019digging}, there is still a huge scope of improvement to bridge the gap between self-supervised and supervised methods. This paper aims to reduce that gap by incorporating novel relational self-attention and data augmentation utilising learnt depth. 

\begin{figure}[t]
\begin{center}
\includegraphics[width=0.8\linewidth]{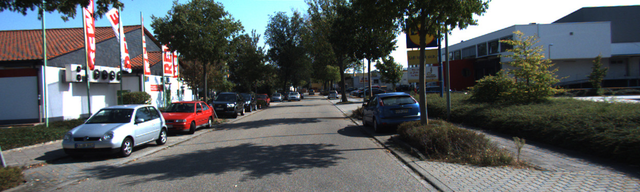}
   \includegraphics[width=0.8\linewidth]{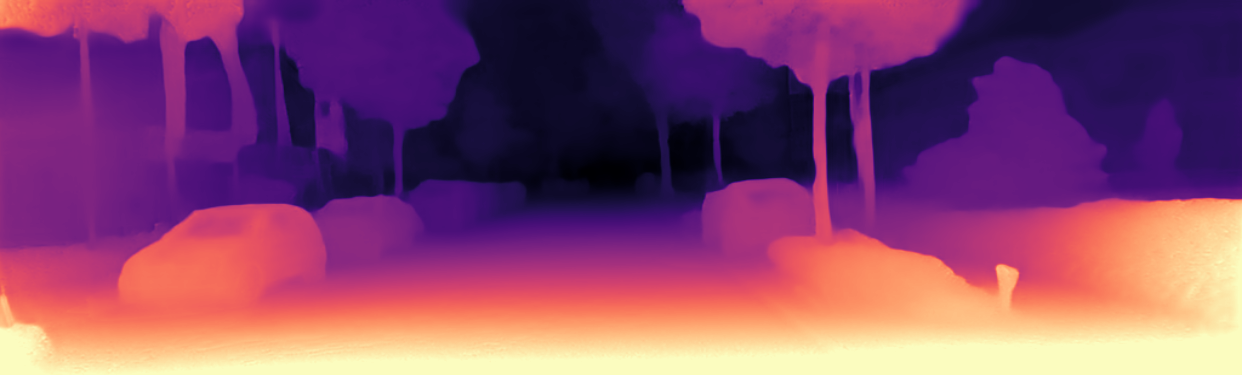}
\end{center}
   \caption{Depth predicted from our network}
\label{fig:onecol}
\end{figure}

We utilise a ResNet18 encoder for our ablation and quantitative analysis and show substantial improvements in learning depth.
Our main contributions are as follows:
\begin{itemize}
    \item We introduce data augmentation as a supervisory loss, improving depth at occluded edges and image boundaries while making the model more robust to illumination changes and image noise. 
    \item Our self attention module learns optimal feature relations that drastically improve our depth prediction.
    \item We show that our novel progressive learning strategy learns robust scale-invariant features leading to significant improvements in depth prediction while saving huge computational overhead of training a high resolution model from scratch.
\end{itemize}
Our network can predict state-of-the-art depth while having significantly lower number of parameters.
\begin{figure*}[ht]
\begin{center}
\includegraphics[width=1\linewidth]{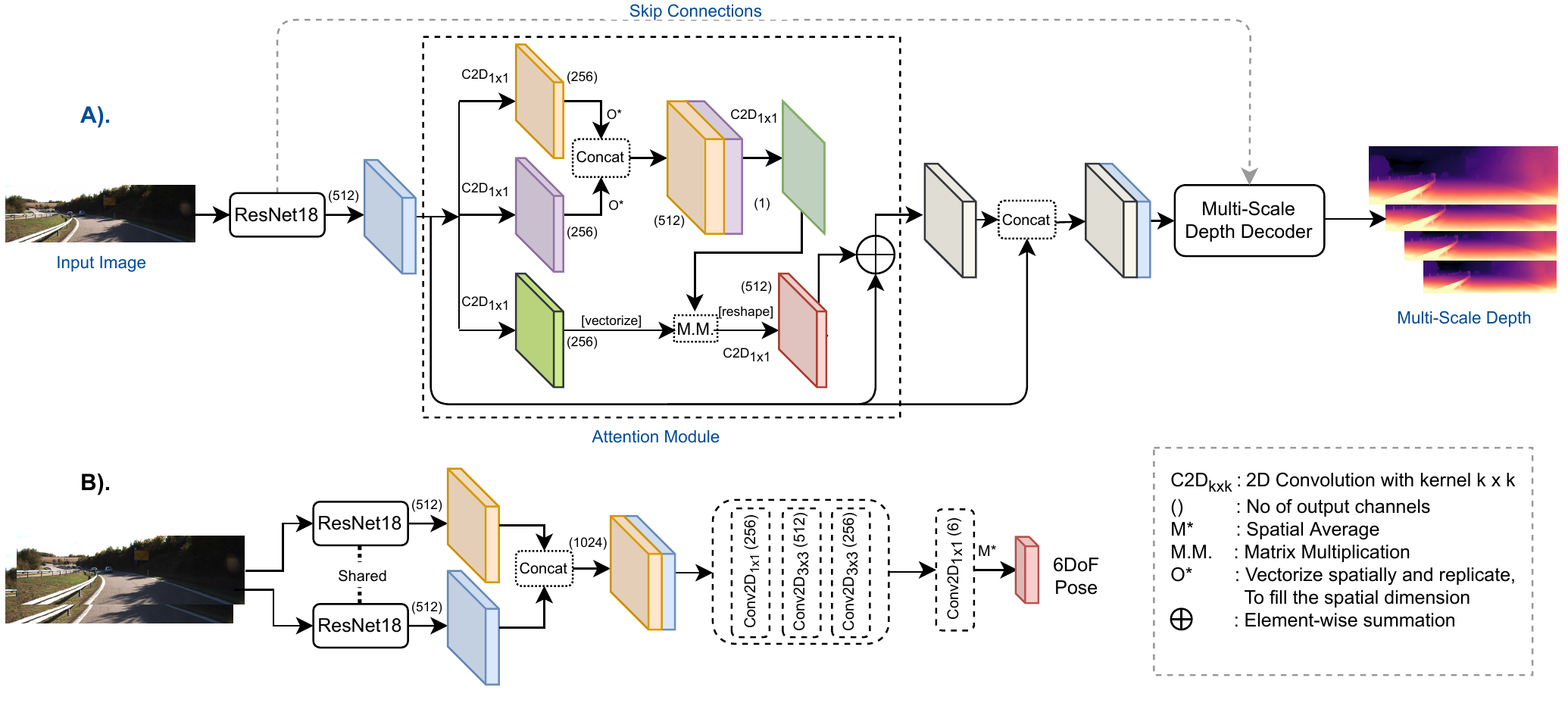}
\end{center}
\setlength{\belowcaptionskip}{-20pt}
\vspace{-15pt}
\caption{Architecture diagram}
\vspace{-15pt}
\label{fig:fig2}
\end{figure*}
\section{Related Work}
Depth estimation from a single colored image is a challenging task due to the obscure nature of this problem. A single depth map can be mapped to innumerable possible colored images. Over the last few years learning models have proven to be successful in effectively learning and exploiting this relationship between color images and their corresponding depths.

\subsection{Supervised Depth Estimation}
Eigen\cite{eigen2014depth} was one of the first ones to explore end to end supervised learning of depth from a single colored image using a multi-scale deep neural network. He trained a model to learn directly from raw colored images and their corresponding depths. Several different approaches have been proposed since then. \cite{10.1109/TPAMI.2008.132} introduced a patch-based model which generated super-pixels to combine local information. \cite{Karsch_2014} used a non-parametric scene sampling pipeline where candidate images from the dataset were matched with target image using high level image and optical flow features. 

Acquiring large amounts of ground truth data in the real world is a challenge and this creates large overheads, both in terms of cost and time as it requires use of lasers like LIDAR. This is the reason that supervised models, despite their superior performance, are not universally applicable. As a result several works have turned to unsupervised or weakly supervised models and use of synthetically generated data.

\cite{wu2018sizetodepth} used real world size of objects to compute depth maps. They used geometric relations to calculate depth maps which were then refined using energy function optimization. \cite{chen2016singleimage} used relative depth annotation instead of actual ground truth depth data. They learned to estimate metric depth using relative depth annotations. These works however, still require supervision signals in the form of additional set of depths or other annotations. Generating large amounts of realistic synthetic data that includes several types of variations found in the real world is not a superficial task as well.
\subsection{Self-supervised Depth Estimation}
A more promising substitute for supervised and weakly supervised models is the self supervised approach. Either stereo or monocular inputs are used for these models. Depth, hallucinated by the model, is used to warp the source image into the target frame. The difference between the reconstructed and reference frame is penalised and added as a reconstruction loss to provide a supervisory signal to the model.

\subsubsection{Self-supervised Stereo Training}
For self-supervised stereo depth estimation, synchronized stereo image pairs are fed into the model.  The model estimates disparity or inverse depth between the two frames and in the process learns to predict the depth of single images. Garg \cite{garg2016unsupervised} presented an approach that reconstructed left images by inverse warping the right images using the predicted depth and known camera extrinsics. The photometric error between the reconstructed image and the original images was used to train the encoder. \cite{godard2017unsupervised} incorporated a left-right consistency term amongst other losses.  \cite{Watson_2019_ICCV} utilised stereo matching to provide sparse supervision in form of depth hints to predict depth. Since then several works have refined self-supervised stereo training of depth. However, some problems still plague stereo estimation. Occlusion drastically affects stereo frames due to the fixed baseline between cameras. Also, wide baseline stereo data might not be available in all real world scenarios e.g. mobile phone camera.

\subsubsection{Self-supervised Monocular Training}
Self supervised monocular depth estimation is naturally unimpeded by a lot of these restraints. In monocular training, temporally consecutive frames are fed into the model instead of stereo pairs. The model has to also learn pose in addition to depth due to the unknown and varying baseline. Zhou et al\cite{zhou2017unsupervised} provided one of the initial works in this domain where they used an end to end learning approach with supervision provided by view synthesis. They used two separate networks for learning depth and pose. \cite{godard2019digging} used a minimum reprojection loss to handle occlusion and prevent the network from learning erroneously from occluded pixels. They computed an automasking framework to prevent learning depth from stationary pixels(static camera). Several works have also incorporated optical flow estimation in their pipelines and tried to exploit relationships between depth, pose and optical flow to achieve more accurate results. \cite{Zou_2018} proposed a cross-task consistency loss, \cite{ranjan2019competitive} performed motion segmentation, \cite{yin2018geonet} decomposed motion into rigid and non rigid components and used a residual flow learning module to handle non rigid cases, \cite{Chen_2019} used losses that ensured 3D structural consistency and enforced geometric constraints,  $S^3$Net\cite{cheng2020s} fuses semantic constraints into depth framework, Shu\cite{shu2020feature} introduces a feature metric loss computed from FeatureNet to improve depth. Huynh \cite{huynh2020guiding} formulates a depth attention volume for guiding monocular depth. Xian \cite{xian2020structure} constructs a structure guided ranking loss for self-supervised learning of depth.
\subsubsection{Self-Attention in Deep learning}
Wang\cite{wang2018non} introduced self-attention as a non-local operation by correlating response at a spatial position as weighted sum of features at all positions. Building on the same framework, Zhang\cite{zhang2019self} utilised self-attention in GANs for image generation tasks. Fu\cite{fu2019dual} formulated a dual attention network for semantic segmentation that unlike traditional works which focus on multi-scale feature fusion, focused on self-attention to integrate local features with their global dependencies adaptively. Since then, self-attention has already been utilised in medical applications\cite{oktay2018attention}, video recognition, semantic segmentation\cite{huang2019ccnet}, object detection\cite{pang2019libra} and video understanding\cite{lin2019tsm}. Unlike a convolutional operation, self-attention provides the ability to learn features and dependencies in non-contiguous regions making it an important building block of deep learning frameworks. We formulate a relational self-attention mechanism, learning from relational reasoning\cite{santoro2017simple} to embed better context in the self-attention framework.
Our model achieves better accuracy without learning for optical flow or motion segmentation by encompassing robust geometric constraints, a relational self-attention framework and utilising augmentation for depth supervision along with our progressive learning strategy.
\addtolength{\headsep}{0.1in}
\begin{figure*}
	\centering
    \resizebox{\textwidth}{!}{
		\input{main_fig/fig.tex}

    }
   \caption{{\bf Qualitative results on the KITTI Eigen split~\cite{eigen2015predicting} test set.} Our models perform better on thinner objects such as trees, signs and bollards, as well as being better at delineating difficult object boundaries. The depth of far objects including sky is further improved. }
    \vspace*{-10pt}
	\label{fig:qa}
\end{figure*}
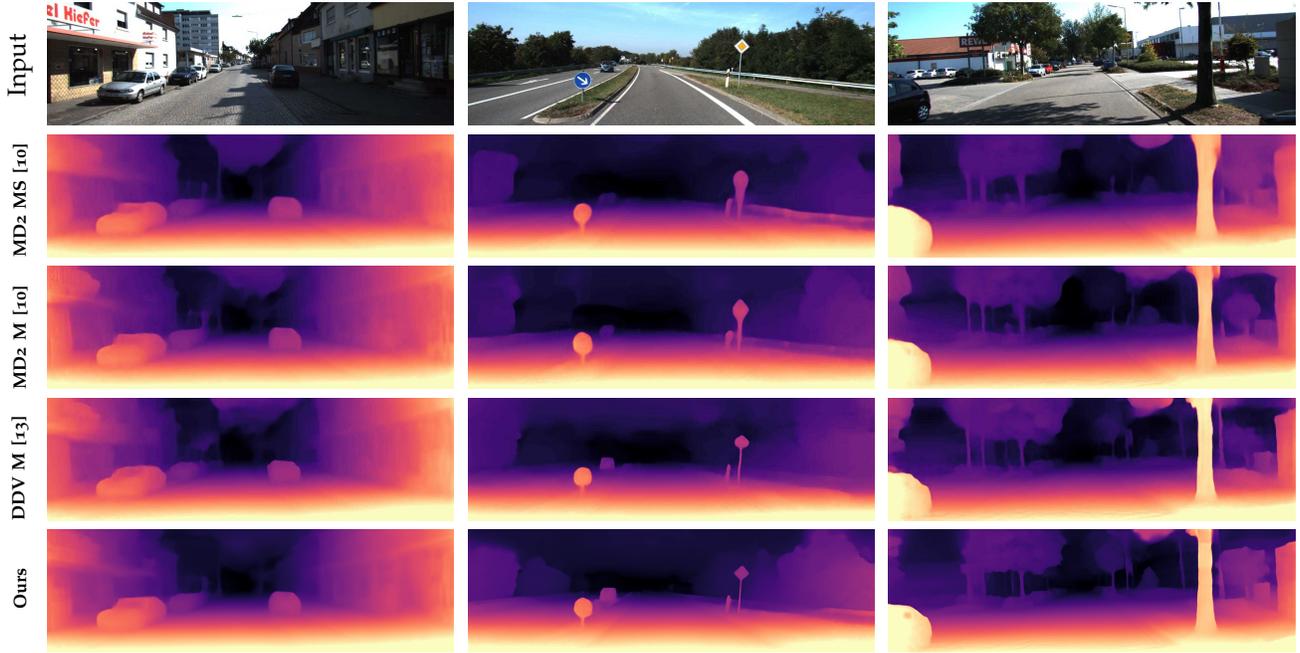

\section{Methodology}
Self-supervised learning utilising photometric consistency has become the de-facto standard for learning depth without ground truth data. The problem of depth prediction is transformed into a problem of view synthesis, where the goal is to use predicted depth of the input image to find per pixel correspondence for reconstructing the input image from another view. By solving for view synthesis, we can train our network to predict depth. We utilise the same approach while incorporating multiple novel data-driven and geometric constraints. Here, we describe a model that jointly learns to predict depth and pose. The network comprises of a shared VGG encoder, depth decoder and pose sub-network. The encoder takes an RGB image as input and extracts it's features that are utilised by both depth decoder and pose sub-network. For training our network we use a 3 frame sequence, where the middle frame is target image $I_t$ and the remaining two frames are source images $I_{S1}$, $I_{S2}$. We predict target depth $D_t$, source depth $D_{S1}$ and $D_{S2}$, pose $P_1$ and pose $P_2$, where pose $P_i$ is the 6DoF transformation from target to the $i^{th}$ source. 

We first outline our training model architecture along with the necessary notations required in formulating losses for training our model then describe in detail the geometric constraints of depth prediction. We describe in detail the augmentation loss framework and the self-attention module for and then delineate each loss along with it's significance in our algorithm.

\subsection{Training Model Architecture}
As shown in \ref{fig:fig2}, our model consists of a ResNet18 encoder \cite{godard2017unsupervised} taking an RGB image as input. Features extracted from the source $I_s$ and target $I_t$ images are concatenated and fed to the pose sub-network to compute the 6x1 egomotion vector. Our depth decoder takes in feature of the target image $I_t$ to predict depth $D_t$ of that image. The encoder-decoder framework is similar to the U-Net architecture introduced by \cite{ronneberger2015u}, that enables us to encapsulate both global as well as local features while predicting depth at 4 scales. The relational attention module takes input as encoder's features and generates attention maps that are concatenated to the original features and fed to the depth decoder as in Figure\ref{fig:fig2}. The pose network comprises of 4 convolutional layers to get a 6x1 output vector\cite{wang2018learning} containing rotation(3x1) and translation (3x1) information as shown in Figure\ref{fig:fig2}. We use Sigmoid activation at depth outputs and ELU activation everywhere else\cite{godard2017unsupervised}. The target image $I_t$ and it's corresponding predicted depth $D_t$ is then processed by the augmentation pipeline to get transformed augmented image $I_{aug}$ and true augmented depth $D_{aug}^{true}$. $I_{aug}$ is then fed to the network to predict output augmented depth $D_{aug}^{out}$. The model returns $D_t, D_s,P ,D_{aug}^{true}, D_{aug}^{out}$ for computing training losses. Target depth $D_t$ warps source image $I_{s}$ to compute synthetic image $I_t'^{s}$ by using bi-linear sampling for sampling source images. While testing, the network can simply compute $D_t$ from $I_t$.

\begin{table*}[ht]
\begin{center}
\begin{tabular}{p{4cm}ccccccc}
\cline{1-8}
Method                             & \cellcolor[HTML]{CBCEFB}Abs Rel & \cellcolor[HTML]{CBCEFB}Sq Rel & \cellcolor[HTML]{CBCEFB}RMSE & \cellcolor[HTML]{CBCEFB}RMSE log & \cellcolor[HTML]{9B9B9B}$\delta$\textless$1.25$ & \cellcolor[HTML]{9B9B9B}$\delta$\textless$1.25^2$ & \cellcolor[HTML]{9B9B9B}$\delta$\textless$1.25^3$                    \\
\cline{1-8}
Zhou et al\cite{zhou2017unsupervised}                      & 0.183                & 1.595                 & 6.709                 & 0.270                 & 0.734                 & 0.902                 & 0.959                 \\
Yang et al\cite{yang2017unsupervised}                        & 0.182                & 1.481                 & 6.501                 & 0.267                 & 0.725                 & 0.906                 & 0.963                 \\
Mahjourian et al\cite{mahjourian2018unsupervised}                  & 0.163                & 1.240                 & 6.220                 & 0.250                 & 0.762                 & 0.916                 & 0.968                 \\
Geonet\cite{yin2018geonet}                         & 0.149                & 1.060                 & 5.567                 & 0.226                 & 0.796                 & 0.935                 & 0.975                 \\
DDVO\cite{wang2018learning}                               & 0.151                & 0.125                & 5.583                 & 0.228                 & 0.81                  & 0.936                 & 0.974                 \\
LEGO\cite{yang2018lego}                                & 0.162                & 1.352                 & 6.276                 & 0.252                 & -                     & -                     & -                     \\
DF-Net\cite{zhan2018unsupervised}                            & 0.150                & 0.124                 & 5.507                 & 0.223                 & 0.806                 & 0.933                 & 0.973                 \\
Ranjan et al\cite{ranjan2019competitive}                     & 0.148                & 0.149                 & 5.464                 & 0.226                 & 0.815                 & 0.935                 & 0.973                 \\
EPC++\cite{luo2018every}                           & 0.141                & 1.029                 & 5.350                 & 0.216                 & 0.816                 & 0.941                 & 0.976                 \\
Struct2Depth(M)\cite{Casser_2019}                    & 0.141               & 1.025                & 5.290                & 0.215                & 0.816                 & 0.945                & 0.979                \\
Monodepth2\cite{godard2019digging}                    & 0.115 & 0.882 & 4.701 & 0.190 & 0.879 & 0.961 & 0.982 \\
DDV\cite{johnston2020self} & 0.106 & 0.861 & 4.699 & 0.185 & 0.889 & 0.962 & 0.982                 \\
\cline{1-8}
\textbf{Proposed Approach} &  \textbf{0.108}  &   \textbf{0.745}  &   \textbf{4.436}  &   \textbf{0.181}  &   \textbf{0.889}  &   \textbf{0.966}  &   \textbf{0.984}  \\
\cline{1-8}
\end{tabular}
\setlength{\belowcaptionskip}{-15pt}
\caption{Self-supervised depth prediction results on KITTI dataset \cite{geiger2013vision} trained at 1024 x 384 resolution. Results on Eigen split \cite{eigen2014depth} for depths at cap 80m, as described in \cite{eigen2014depth}.}
\vspace{-25pt}
\label{table:tab1}
\end{center}
\end{table*}

\subsection{Constraints for depth prediction}
In this section, we describe the formulation of various loss functions used in our network for self-supervised learning of depth and pose. 

\subsubsection{Minimum Photometric Loss}
As described by \cite{godard2019digging}, this loss is a slight variation from the normal photometric loss. Instead of taking per pixel average of photometric loss for all sources, we compute minimum of photometric loss for all sources. This successfully tackles scenarios where a target pixel is visible in one source image but not visible in the other source image due to occlusion and only back-propagates the minimum error, thereby ignoring the erroneous one.
\begin{equation}
    L_{p} = \min\limits_{{s}\in(1,2)} pe(I_{t},{I'}_{t}^{s}) 
\end{equation}
Here, photometric error $pe$ is defined by a weighted combination of  L1 loss and Structural Similarity (SSIM) \cite{wang2004image}, similar to \cite{godard2019digging}\cite{Chen_2019}\cite{ranjan2019competitive}.
\begin{equation}
    pe(I_t,I_s) = \frac{\alpha}{2}  (1-SSIM(I_t,I_s)) + (1-\alpha) ||I_t-I_s|| 
\end{equation}
,where $\alpha=0.85$
Similar to \cite{godard2019digging}, we apply a per pixel binary mask $\lambda$ to the computed losses. The mask $\lambda$ is generated by comparing the photometric error between source and target frames with that between the synthesised source and target frames.
\begin{equation}\label{eq:eq5}
    \lambda = \![ \min\limits_{s} pe(I_t,{I'}_{t}^s) < \min\limits_{s} pe(I_t,I_s)   \!]
\end{equation}
This eliminates static pixels from corrupting the loss and the network skips learning depth altogether if the camera isn't moving.  
We observe that although this improves depth prediction drastically, it leads to random white noise around static regions and makes the learning of depth more sensitive to noisy images.  
This happens because the mask doesn't consider neighbouring pixels while comparing photometric errors and simply takes a threshold of per pixel values. 
To alleviate this problem, we enforce a L1 regularisation over inverse of $\lambda$, thereby motivating the mask to be positive for those sparse number of pixels.

\begin{equation}
    L_r = |1 - \lambda|
\end{equation}
We compute first order gradient smoothness loss $L_s$\cite{godard2017unsupervised} over mean normalized inverse depth $d_t$\cite{wang2018learning} to ensure that the predicted depth is locally smooth as well as consistent in textured regions.
\begin{equation}
\begin{split}
    L_{s} = \abs{\delta_{x}d_t} e^{-\abs{\delta_{x}I_t}} + \abs{\delta_{y}d_t} e^{-\abs{\delta_{y}I_t}}
\end{split}
\end{equation}
where $\delta_{x}$ and $\delta_{y}$ are gradients in horizontal and vertical direction respectively.

\subsubsection{Data augmentation for depth supervision}
Several works have utilised data augmentation\cite{godard2019digging,pillai2019superdepth,johnston2020self} in their deep learning pipeline to make their networks more robust to challenging scenarios and invariant to changes in noise, brightness and contrast that are common in the real world. Traditionally, pipelines performed data augmentation at the data loading stage and the augmented data was fed into the network for training. We utilise augmentation for generating augmented inputs and outputs that are used to train the network in a semi-supervised manner. We incorporate an augmentation loss, in the form of a depth supervision, that improves the predicted depth.
While training, in the first forward pass, the network takes $I_t$ as input, giving depth $D_t$ as network output. We pass the pair $(I_t,D_t)$ to the augmentation pipeline, applying identical random image cropping, flipping, skewing and scaling and affine transformations to both. Additionally, we perform random changes to brightness, jitter, gamma and saturation to the input image $I_t$. We also add random gaussian noise to $I_t$. The augmentation pipeline returns augmented image $I_{aug}$ and true augmented depth $D^{true}_{aug}$. While training, in the second forward pass, $I_{aug}$ is fed to the network generating augmented predicted depth $D^{out}_{aug}$. Augmented depth maps generated in first pass serve as ground truth for depth maps generated in the second pass. The augmentation loss minimises the difference between the output augmented depth and the true augmented depth, enforcing both depths to be consistent with each other. 
\begin{equation}
L_{a} = ||D_{aug}^{true} - D_{aug}^{out}||_1
\end{equation}
Due to camera egomotion, occlusion is present at certain image boundaries. Rescaling and crop transformations randomly remove boundary regions from the image, while ensuring that it's size remains the same. Thus, the boundaries of the augmented depth are more accurate due to lower probability of occlusion.
\subsubsection{Relational Self-Attention}
The relational self-attention block takes input the features $X$ from the ResNet18 encoder and computes self-attention $Y$ that is added as a residual connection to the input feature $X$ to compute the output feature. The operation can be summarised as follows:
\begin{equation}
    y_{i} = \frac{1}{N} \sum_{\forall{j}} W_{f}^{T}[\theta(x_{i}), \phi(x_{j})] g(x_{j})
\end{equation}

Here, $W_{f}$ is a weight factor that projects the concatenated vector to the scalar by performing a convolution with single output channel, $[.,.]$ denotes concatenation, $N$ defines the number of positions in $X$. 
Also, the functions $\theta$, $\phi$ and $g$ are defined by $1x1$ 2D convolution operations as shown in Figure \ref{fig:fig2}. The input $X$ generates the projection, query, key and value embeddings as
\begin{equation}
\begin{split}
    f(\mathbf{X}) = & \mathbf{W}_f\mathbf{X}, \\ g(\mathbf{X}) = & \mathbf{W}_g\mathbf{X}, \\
    \theta(\mathbf{X}) = & \mathbf{W}_{\theta}\mathbf{X}, \\
    \phi(\mathbf{X}) = & \mathbf{W}_{\phi}\mathbf{X},
\end{split}
\end{equation}
where $W_f$, $W_g$, $W_\theta$, $W_{\phi} \in \mathbb R^{2}$ are weight matrices to be learnt. 
The pairwise relation between query $\theta$ and key $\phi$ is projected by $W_f$ and multiplied by value $g(X)$ to compute our relational self-attention that is then element-wise added to input $X$ to give output of our attention block. 

The output is then concatenated with the encoder's features and utilised by the decoder to computed multi-scale depth.

\subsubsection{Final Training Loss} 
We combine photometric and smoothness losses with our data augmentation loss along with $L_1$ regularization over mask to obtain our final objective. 
\begin{equation}
   L_{f}=   \alpha_p L_{p} + \alpha_s L_{s} +  \alpha_a L_{a} + \alpha_r L_{r} 
\end{equation}
All our losses are computed per-pixel and averaged over entire image, scales and batch. 

\begin{table}[ht]
\begin{center}
    \resizebox{\linewidth}{!}{
      \begin{tabular}{|l|c||c|c|c|c|}
      \hline
       & Type & \cellcolor{col1}Abs Rel & \cellcolor{col1}Sq Rel  & \cellcolor{col1}RMSE & \cellcolor{col1}$\text{log}_{10}$ \\
      \hline
      Karsch \cite{karsch2014depth} & D & 0.428 & 5.079 & 8.389 & 0.149 \\
      Liu \cite{liu2014discrete}& D & 0.475 & 6.562 & 10.05 & 0.165 \\
      Laina \cite{laina2016deeper}& D & {\bf 0.204} & {\bf 1.840} & {\bf 5.683} & {\bf 0.084} \\ \hline
      Monodepth \cite{godard2017unsupervised} & S & 0.544 & 10.94 & 11.760 & 0.193 \\
      Zhou \cite{zhou2017unsupervised} & M &  0.383 & 5.321 & 10.470 & 0.478 \\
      DDVO \cite{wang2017learning} & M & 0.387 & 4.720 & 8.090 & 0.204 \\
      Monodepth2 \cite{godard2019digging} & M & 0.322 & 3.589 & 7.417 & 0.163  \\
      DDV \cite{johnston2020self} & M & 0.297 & 2.902 & 7.013 & 0.158 \\ 
      \textbf{Proposed Approach} & \textbf{M} & \textbf{0.289} & \textbf{2.552} & \textbf{6.869} & \textbf{0.155} \\ %
      \hline
      \end{tabular}
    }
    \end{center}  
    \vspace{-2pt}
    \caption{{\bf Make3D\cite{saxena2009make3d} results.} All self-supervised monocular (M) methods use median scaling.}
    \vspace{-22pt}
\label{table:tab2}
\end{table}

\section{Experiments and Results}
This section introduces the dataset and describes the training details. We describe in detail various comparative qualitative and quantitative studies along with an ablation study undertaken for validation and show that our method surpasses all other existing related methods.

\begin{table*}[t]
\begin{center}
\begin{tabular}{p{0.8cm}cccccccc}
\cline{1-9}
Aug Loss & Attention & \cellcolor[HTML]{CBCEFB}Abs Rel & \cellcolor[HTML]{CBCEFB}Sq Rel & \cellcolor[HTML]{CBCEFB}RMSE & \cellcolor[HTML]{CBCEFB}RMSE log & \cellcolor[HTML]{9B9B9B}$\delta$\textless$1.25$ & \cellcolor[HTML]{9B9B9B}$\delta$\textless$1.25^2$ & \cellcolor[HTML]{9B9B9B}$\delta$\textless$1.25^3$
\\
\cline{1-9}
No                  & No              & 0.115  &   0.919  &   4.854  &   0.194  &   0.877  &   0.958  &   0.980  \\
Yes                        & No                      &  0.113  &   0.837  &   4.726  &   0.189  &   0.879  &   0.961  &   0.982  \\
No                           & Yes                      & 0.111  &   0.827  &   4.742  &   0.189  &   0.878  &   0.960  &   0.982  \\ 
\cline{1-9}
\textbf{Yes} & \textbf{Yes} & \textbf{0.111}  &   \textbf{0.817}  &   \textbf{4.685}  &   \textbf{0.188}  &   \textbf{0.883}  &   \textbf{0.961}  &   \textbf{0.982}
\\\cline{1-9}
\end{tabular}
\end{center}
\setlength{\belowcaptionskip}{-10pt}
\caption{Ablation study for depth prediction at 640x192 image resolution using ResNet18 Encoder on Eigen split\cite{eigen2014depth}. We observe that the combination of Augmentation Loss and our Attention framework gives us the best depth results.}
\label{table:tab3}
\vspace*{-10pt}
\end{table*}

\begin{table*}[ht]
\begin{center}
\begin{tabular}{l|c|c|c|c|c|c|c|c}
\cline{1-9}
                            Method & Backbone & \cellcolor[HTML]{CBCEFB}Abs Rel & \cellcolor[HTML]{CBCEFB}Sq Rel & \cellcolor[HTML]{CBCEFB}RMSE & \cellcolor[HTML]{CBCEFB}RMSE log & \cellcolor[HTML]{9B9B9B}$\delta$\textless$1.25$ & \cellcolor[HTML]{9B9B9B}$\delta$\textless$1.25^2$ & \cellcolor[HTML]{9B9B9B}$\delta$\textless$125^3$                        \\             \cline{1-9}
Monodepth2 \cite{godard2019digging} & ResNet18      & 0.115 & 0.902 & 4.847 & 0.193 & 0.877 & 0.960 & 0.981                        \\
DDV \cite{johnston2020self} & ResNet18 & 0.111 & 0.941 & 4.817 & 0.189 & 0.885 & 0.961 & 0.981
\\ \cline{1-9}
\textbf{Proposed Approach} & \textbf{ResNet18} & \textbf{0.111}  &  \textbf{0.817}  &  \textbf{4.685}  &   \textbf{0.188}  &   \textbf{0.883}  &   \textbf{0.961}  &   \textbf{0.982} \\
\textbf{Proposed Approach 1024x384} &  \textbf{ResNet18} & \textbf{0.108}  &   \textbf{0.745}  &   \textbf{4.436}  &   \textbf{0.181}  &   \textbf{0.889}  &   \textbf{0.966}  &   \textbf{0.984}  \\
\cline{1-9}
\end{tabular}
\end{center}
\setlength{\belowcaptionskip}{-10pt}
\caption{Comparing our method at 640x192 resolution with other methods utilising same network backbone.}
\label{table:tab4}
\vspace*{-10pt}
\end{table*}

\begin{table*}[ht]
\begin{center}
\begin{tabular}{l|c|c|c|c|c|c|c}
\cline{1-8}
                            Depth Decoder's Input & \cellcolor[HTML]{CBCEFB}Abs Rel & \cellcolor[HTML]{CBCEFB}Sq Rel & \cellcolor[HTML]{CBCEFB}RMSE & \cellcolor[HTML]{CBCEFB}RMSE log & \cellcolor[HTML]{9B9B9B}$\delta$\textless$1.25$ & \cellcolor[HTML]{9B9B9B}$\delta$\textless$1.25^2$ & \cellcolor[HTML]{9B9B9B}$\delta$\textless$125^3$                        \\             \cline{1-8}
Attention + No feature concat &  0.113  &   0.879  &   4.777  &   0.190  &   0.880  &   0.959  &   0.981  \\
Attention in all skip connections + Feature concat &   0.112  &   0.856  &   4.699  &   0.188  &   0.880  &   0.961  &   0.982  \\
Attention + feature concat in all skip connections &   0.112  &   0.866  &   4.742  &   0.189  &   0.879  &   0.960  &   0.982 \\
\textbf{Attention + Feature concat} & \textbf{0.111}  &  \textbf{0.817}  &  \textbf{4.685}  &   \textbf{0.188}  &   \textbf{0.883}  &   \textbf{0.961}  &   \textbf{0.982} \\
\cline{1-8}
\end{tabular}
\end{center}
\setlength{\belowcaptionskip}{-10pt}
\caption{Comparing our method at 640x192 resolution with multiple variations of self-attention features. Augmentation loss is applied during training. Feature is the ResNet18 encoder's output feature.}
\label{table:tab5}
\vspace*{-15pt}
\end{table*}

\subsection{Dataset}  Our model was trained on KITTI 2015 dataset\cite{geiger2013vision}. This dataset comprises videos captured by a camera mounted on a car moving through the German city of Karlsruhe and is widely recognized and often used for tasks like estimation of depth, optical flow and car's egomotion. We used the Eigen test split\cite{eigen2014depth} of this dataset and tested our model using the ground truth labels present in it. The test set consists of 697 images and it is ensured that frames that are similar to those present in the test set are removed from the training set. We also test our trained model on the 134 images in Make3D dataset\cite{saxena2009make3d}.
\subsection{Parameter Settings} Similar to other self-supervised models\cite{godard2019digging}\cite{ranjan2019competitive}\cite{Chen_2019}, we use ImageNet weights for initialising our network and train our model using a single NVIDIA 2080Ti GPU. Three temporally consecutive images are fed into the model and Adam optimizer with $\beta_1 = 0.9$ and $\beta_2 = 0.999$ is used. Initial learning rate is set to $10^{-4}$ and batch size to 12. While optimizing our network we set weights of different loss terms to $\alpha_p=1.0, \alpha_s=0.001, \alpha_a=0.1$ and $\alpha_r=0.001$. While preparing training data, static frames are removed from the dataset as proposed by Zhou et al\cite{zhou2017unsupervised}. Basic augmentation in the form of random cropping, color jittering, resizing and flipping is also performed as part of our data preparation pipeline. We train our network over two phases in a progressive manner. In the first phase, images of 640x192 resolution are fed into the network. After training it for 50 epochs, in the second phase, we freeze the pose encoder and feed the higher resolution 1080x384 images to the depth network and train our model for 5 epochs with batch size 2. Progressive training aids in further improvement and faster convergence in our depth prediction model at a higher resolution as shown in Table \ref{table:tab4}.

\subsection{Main Results} We compare our results with other recent models in Table \ref{table:tab1}. These results show that our monocular model is able to comprehensively outperform all existing state of the art self-supervised monocular methods. Our model is even able to surpass methods that incorporate optical flow prediction into their pipeline\cite{yin2018geonet}, \cite{Chen_2019},\cite{ranjan2019competitive}, while having lower number training parameters. During evaluation, as common practice\cite{eigen2014depth}, we cap depth to 80m. Table \ref{table:tab4} shows the comparison with DDV\cite{johnston2020self} and moonodepth2\cite{godard2019digging} trained using features from same ResNet18 encoder. Our method has better RMSE, Sq Rel, Abs Rel than other similar methods. This shows that our model performs significantly better in all metrics on Eigen split of KITTI 2015 dataset. 

\subsection{Qualitative Analysis}
Figure \ref{fig:qa} displays qualitative improvements in our method over baseline Monodepth2(MD2)\cite{godard2019digging} and DDV\cite{johnston2020self}. Our algorithm retains structural details in objects like poles, sign boards and trees while learning smooth depth over entire scene. We also have the least noise in disparity values of the infinitely distant sky. 

\subsection{Make3D}
Table \ref{table:tab2} shows results of our model that is trained on KITTI dataset and tested on the Make3D dataset\cite{saxena2009make3d}. We use the crop defined by \cite{zhou2017unsupervised} and apply depth median scaling for fair comparison. The table shows our method's superior performance than other self-supervised methods while bridging the gap between supervised ones\cite{laina2016deeper}.

\subsection{Ablation Study of Losses}
We also undertake exhaustive quantitative comparison of all the losses to analyze the impact of each loss component. Table \ref{table:tab3} shows different combinations of losses applied and the corresponding results achieved by our model. It is evident from the table, that with just the inclusion of augmentation loss, we get significant gains over the baseline. The augmentation loss makes the model more robust to variation in brightness, contrast and image noise. Supervising the network in form of augmentation loss utilising the true augmented depth drastically improves the depth prediction at occluded regions including image boundaries. 
Similarly, appearance and color based transforms help the network in learning to predict more consistent and robust depth which is less affected by noise and illumination changes.
We observe that adding reflection padding to our network doesn't have noticeable effect on the depth prediction results as the augmentation loss already improves depth at image boundaries. We also observe that attention improves $Abs Rel$ more than Augmentation and the combination of both losses have a multi-fold improvement over baseline. We also tried replacing skip connections by attention module but the added complexity was drastically high with no significant improvement in depth prediction. As depicted in Table \ref{table:tab5}, concatenating encoder feature to the attention gave better results than simply passing attention block's output to the decoder. This tells us that attention though significant isn't sufficient to achieve optimal result.  Also, increasing the augmentation loss weight $\alpha_a > 0.1$ induced texture copy artifacts and decreasing it led to minimal improvement in accuracy.

As observed by Monodepth2\cite{godard2019digging}, it is necessary to handle static frames, i.e. frames where either the camera is stationary or regions such as sky that does not change across consecutive frames. Automasking masks out these areas and prevents the model from learning erroneous depth. To enforce mask to be consistent and smooth, and eliminating noisy values, we apply L1 regularisation over inverse of our mask. This slightly increases the number pixels to be evaluated and reduces artifacts over still regions like sky.
Our methods predicts superior results both qualitatively and quantitatively when compared to other self-supervised monocular depth prediction methods.

\section{Conclusion}
We propose a self-supervised model which utilises relational self-attention for jointly learning depth and camera egomotion. The model is able to predict accurate and sharp depth estimates by incorporating data augmentation as depth supervision.
Our algorithm predicts state-of-the-art depth on the KITTI bench-mark\cite{geiger2013vision}.  We achieve  In future, we shall utilise optical flow for motion segmentation, pretrained models and semantic cues for further strengthening depth of moving objects. Architectural innovations in deep learning such as vision transformers along with cues like optical flow and semantic information present in the scene can further optimize robustness and consistency in predicting depth.

\bibliographystyle{root}
\bibliography{root}

\end{document}

%% file: main_fig/fig.tex
\newcommand{\turnheightnew}{0.195\columnwidth}
\centering
\begin{tabular}{@{\hskip 2mm}c@{\hskip 2mm}c@{\hskip 2mm}c@{\hskip 2mm}c@{}}
{\rotatebox{90}{\hspace{4mm}Input}} &
\includegraphics[height=\turnheightnew]{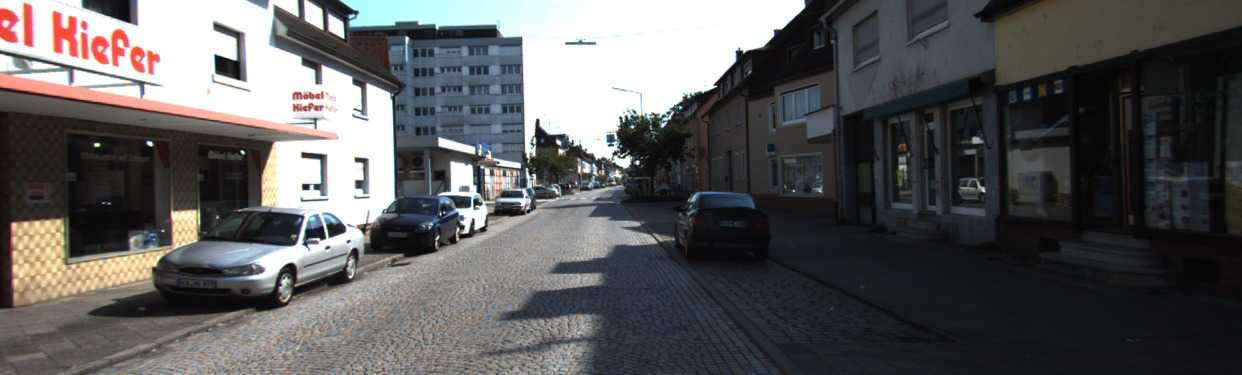} &
\includegraphics[height=\turnheightnew]{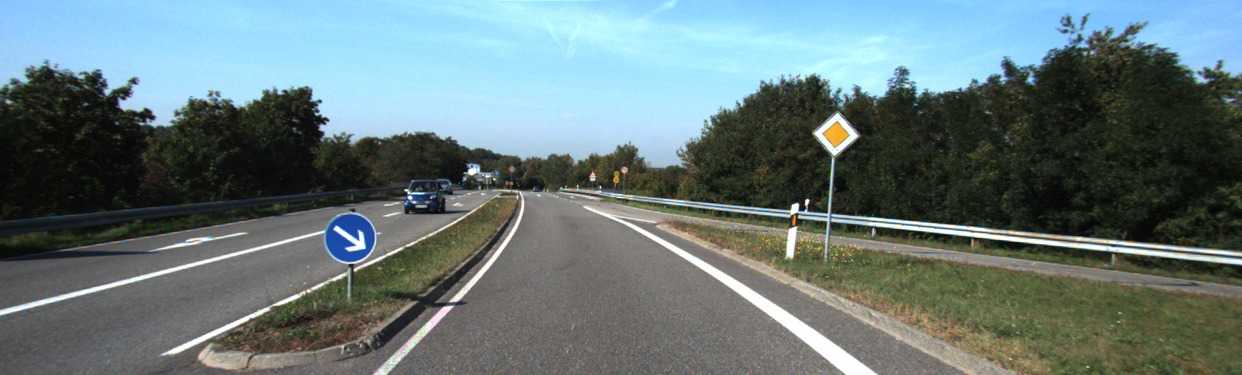} &
\includegraphics[height=\turnheightnew]{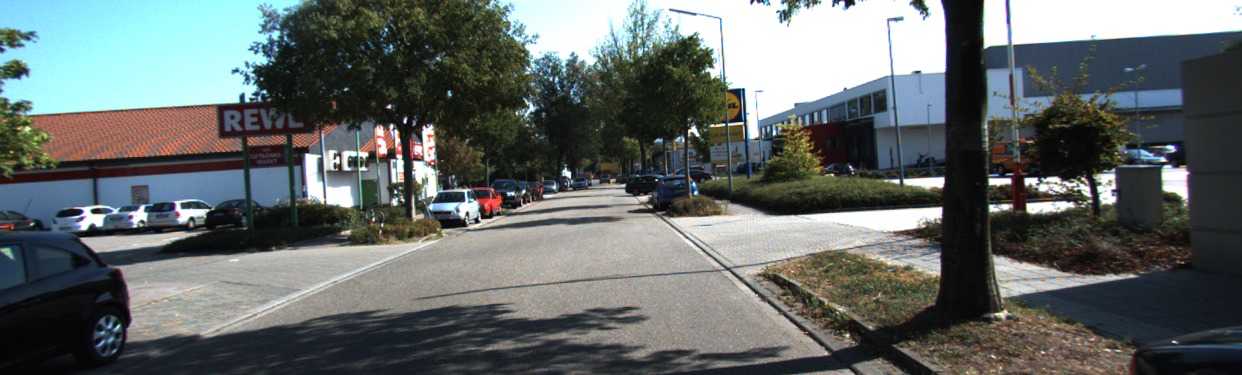} \\

{\rotatebox{90}{\scriptsize \hspace{0mm}\textbf{MD2 MS~\cite{godard2019digging}}}} &
\includegraphics[height=\turnheightnew]{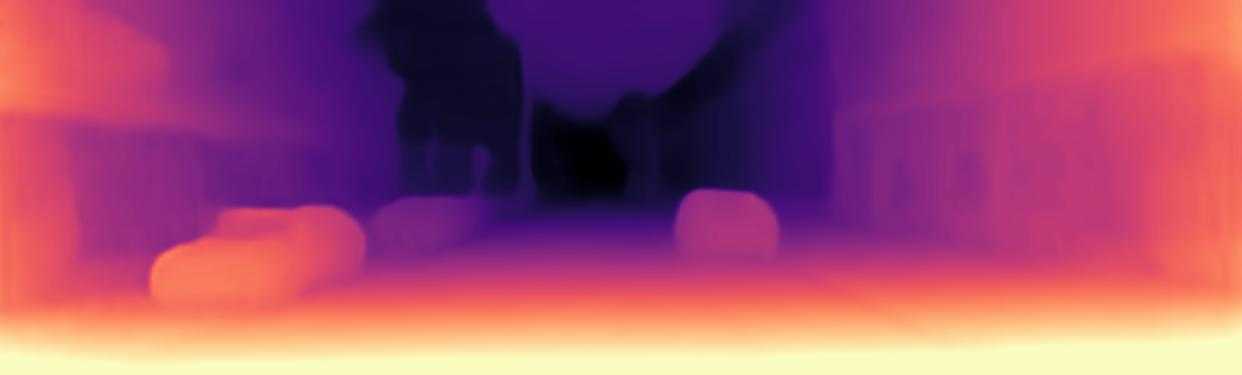} &
\includegraphics[height=\turnheightnew]{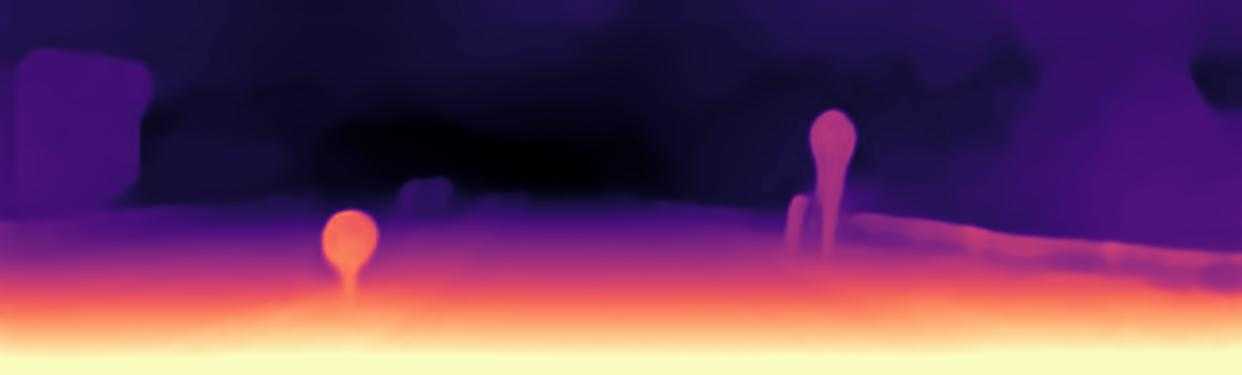} &
\includegraphics[height=\turnheightnew]{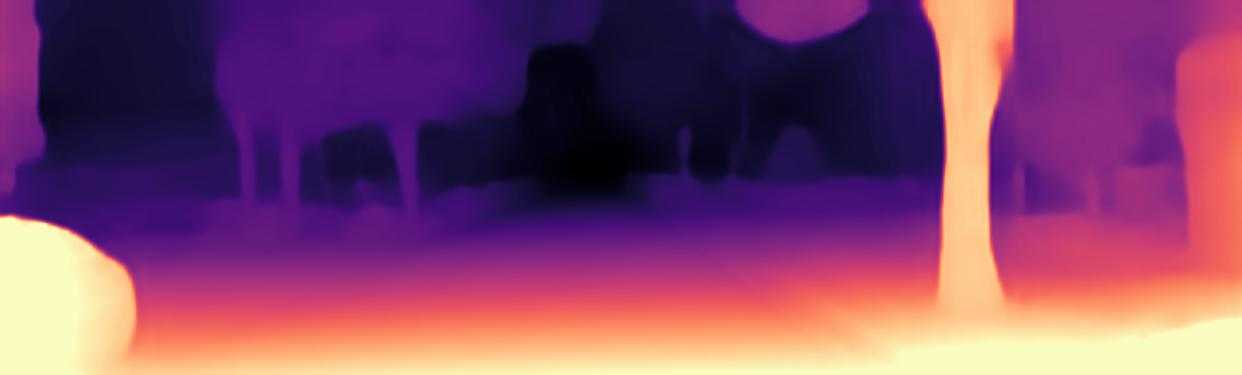} \\

{\rotatebox{90}{\scriptsize \hspace{0mm}\textbf{MD2 M~\cite{godard2019digging}}}} &
\includegraphics[height=\turnheightnew]{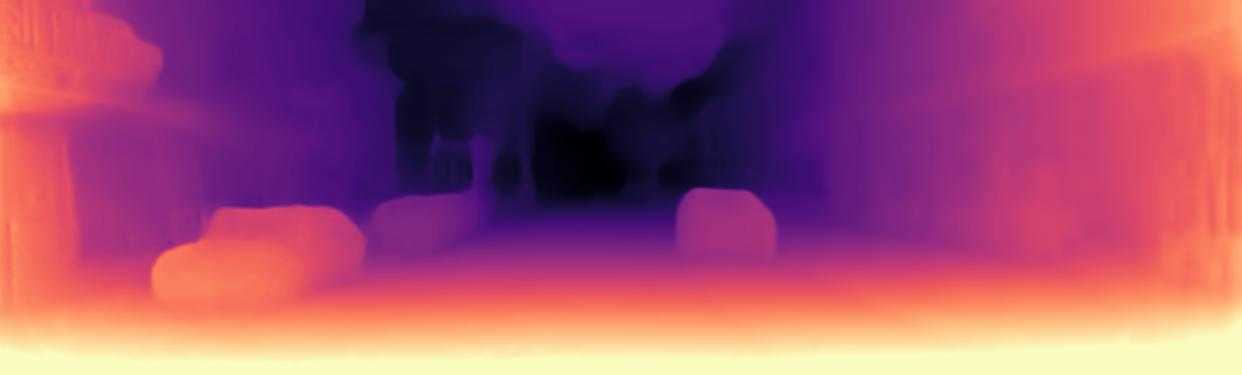} &
\includegraphics[height=\turnheightnew]{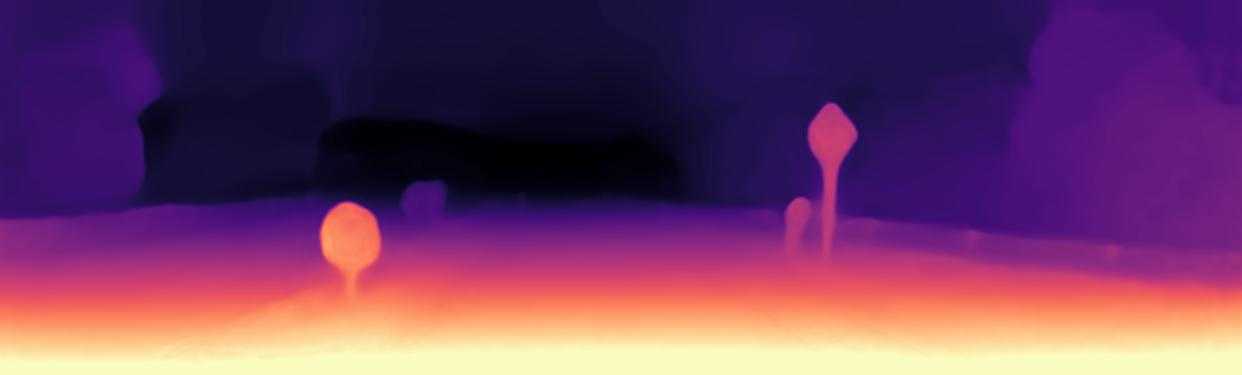} &
\includegraphics[height=\turnheightnew]{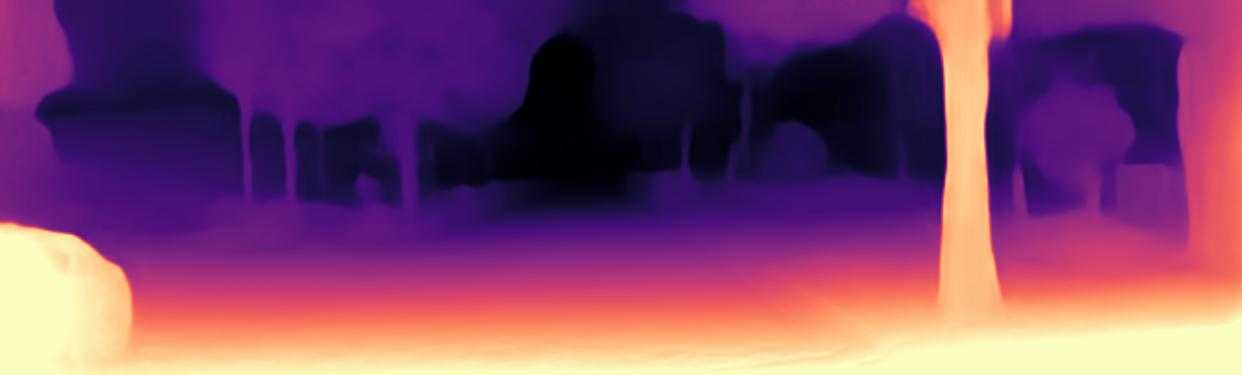} \\

{\rotatebox{90}{\scriptsize \hspace{0mm}\textbf{DDV M~\cite{johnston2020self}}}} &
\includegraphics[height=\turnheightnew]{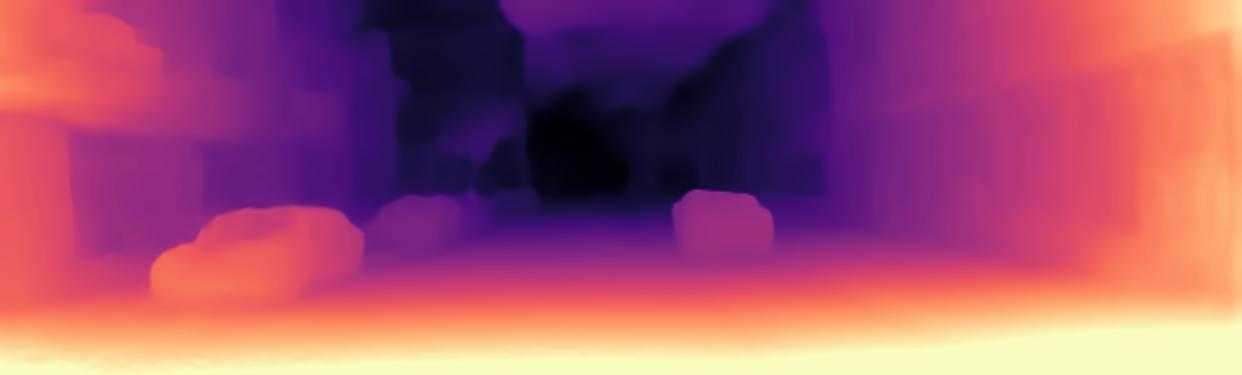} &
\includegraphics[height=\turnheightnew]{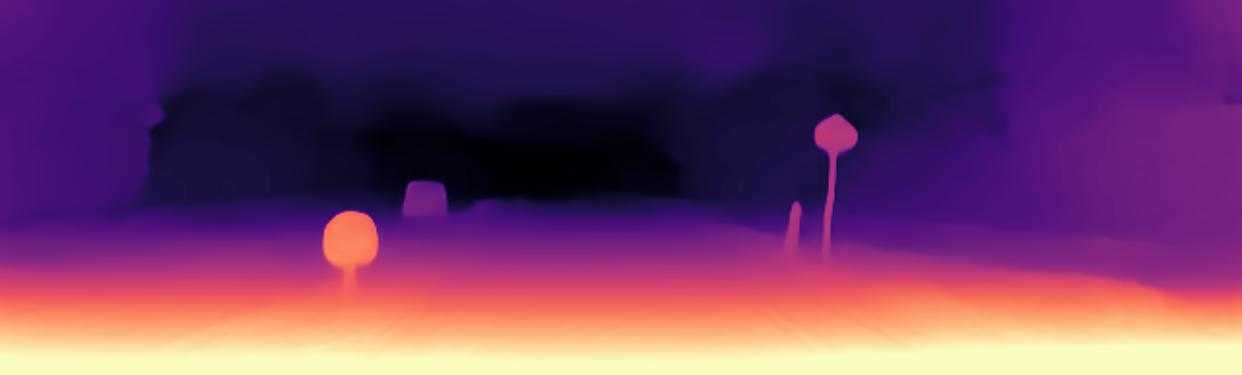} &
\includegraphics[height=\turnheightnew]{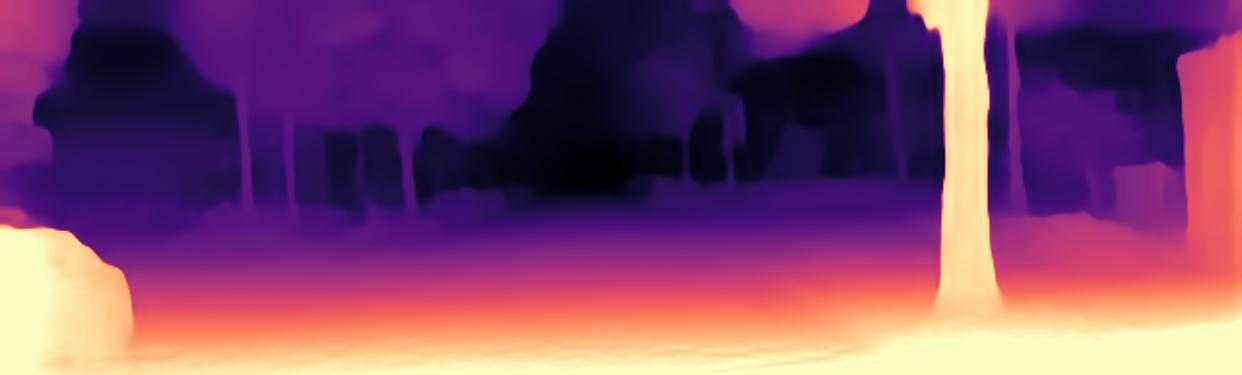} \\

{\rotatebox{90}{\scriptsize \hspace{6mm}\textbf{Ours}}} &
\includegraphics[height=\turnheightnew]{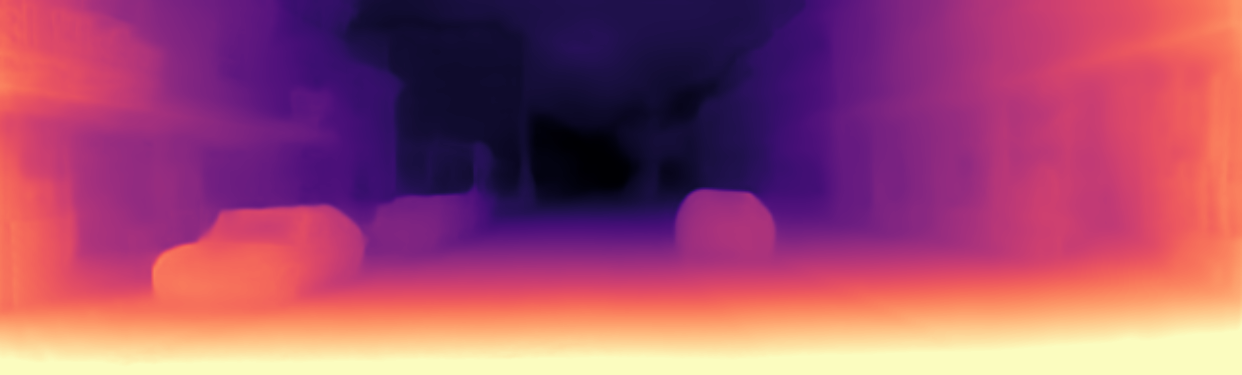} &
\includegraphics[height=\turnheightnew]{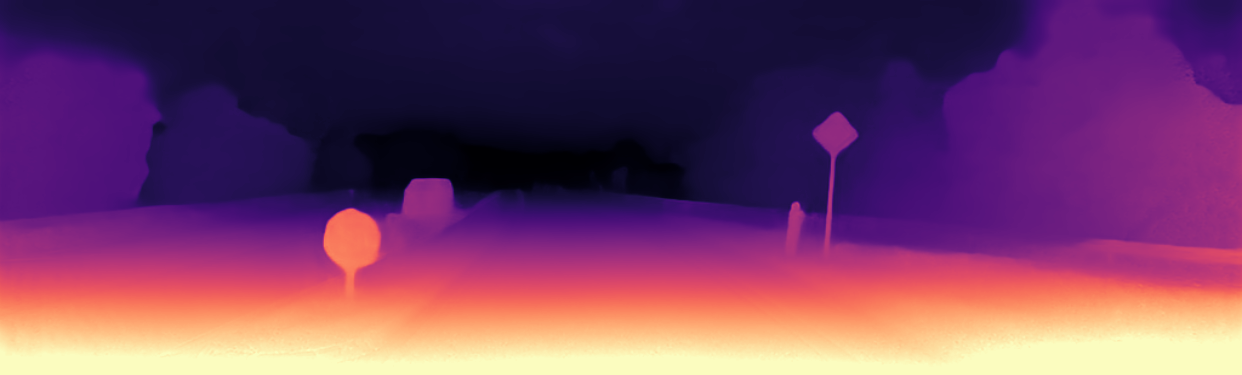} &
\includegraphics[height=\turnheightnew]{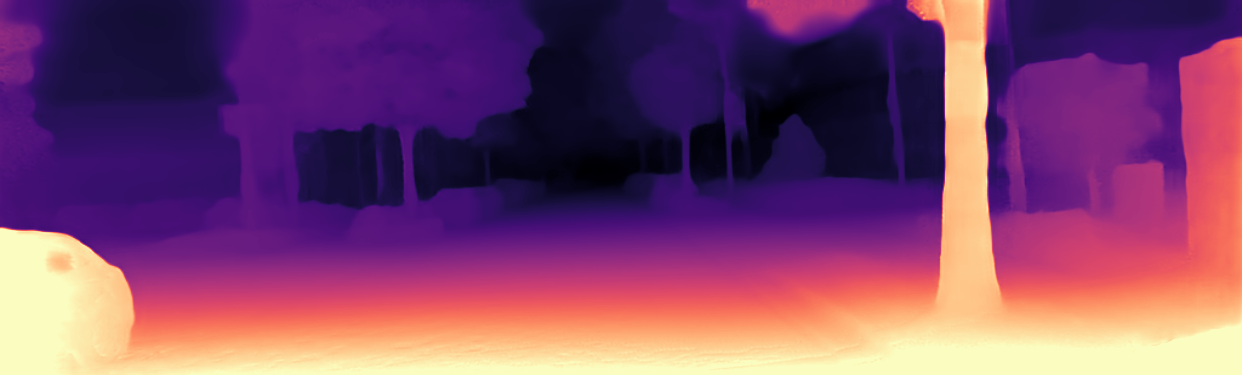} \\
\end{tabular}

%% file: root.bbl
\begin{thebibliography}{10}
\providecommand{\url}[1]{#1}
\csname url@rmstyle\endcsname
\providecommand{\newblock}{\relax}
\providecommand{\bibinfo}[2]{#2}
\providecommand\BIBentrySTDinterwordspacing{\spaceskip=0pt\relax}
\providecommand\BIBentryALTinterwordstretchfactor{4}
\providecommand\BIBentryALTinterwordspacing{\spaceskip=\fontdimen2\font plus
\BIBentryALTinterwordstretchfactor\fontdimen3\font minus
  \fontdimen4\font\relax}
\providecommand\BIBforeignlanguage[2]{{%
\expandafter\ifx\csname l@#1\endcsname\relax
\typeout{** WARNING: IEEEtran.bst: No hyphenation pattern has been}%
\typeout{** loaded for the language `#1'. Using the pattern for}%
\typeout{** the default language instead.}%
\else
\language=\csname l@#1\endcsname
\fi
#2}}

\bibitem{Mauri2020DeepLF}
A.~Mauri, R.~Khemmar, B.~Decoux, N.~Ragot, R.~Rossi, R.~Trabelsi, R.~Boutteau,
  J.-Y. Ertaud, and X.~Savatier, ``Deep learning for real-time 3d multi-object
  detection, localisation, and tracking: Application to smart mobility,''
  \emph{Sensors (Basel, Switzerland)}, vol.~20, 2020.

\bibitem{Valentin2018DepthFM}
J.~P.~C. Valentin, A.~Kowdle, J.~T. Barron, N.~Wadhwa, M.~Dzitsiuk,
  M.~Schoenberg, V.~Verma, A.~Csaszar, E.~Turner, I.~Dryanovski, J.~Afonso,
  J.~Pascoal, K.~Tsotsos, M.~Leung, M.~Schmidt, O.~G. Guleryuz, S.~Khamis,
  V.~Tankovich, S.~R. Fanello, S.~Izadi, and C.~Rhemann, ``Depth from motion
  for smartphone ar,'' \emph{ACM Transactions on Graphics (TOG)}, vol.~37, pp.
  1 -- 19, 2018.

\bibitem{Bello2019Deep3N}
J.~L.~G. Bello and M.~Kim, ``Deep 3d-zoom net: Unsupervised learning of
  photo-realistic 3d-zoom,'' \emph{ArXiv}, vol. abs/1909.09349, 2019.

\bibitem{Liu2019AuroraGR}
Y.~Liu, Y.~Tai, J.-L. Li, S.~Ding, C.~Wang, F.~Huang, D.~Li, W.~Qi, and R.~Ji,
  ``Aurora guard: Real-time face anti-spoofing via light reflection,''
  \emph{ArXiv}, vol. abs/1902.10311, 2019.

\bibitem{Kim2020ImageDM}
Y.-J. Kim and C.~Yim, ``Image dehaze method using depth map estimation network
  based on atmospheric scattering model,'' \emph{2020 International Conference
  on Electronics, Information, and Communication (ICEIC)}, pp. 1--3, 2020.

\bibitem{eigen2014depth}
D.~Eigen, C.~Puhrsch, and R.~Fergus, ``Depth map prediction from a single image
  using a multi-scale deep network,'' in \emph{Advances in neural information
  processing systems}, 2014, pp. 2366--2374.

\bibitem{fu2018deep}
H.~Fu, M.~Gong, C.~Wang, K.~Batmanghelich, and D.~Tao, ``Deep ordinal
  regression network for monocular depth estimation,'' in \emph{Proceedings of
  the IEEE Conference on Computer Vision and Pattern Recognition}, 2018, pp.
  2002--2011.

\bibitem{godard2017unsupervised}
C.~Godard, O.~Mac~Aodha, and G.~J. Brostow, ``Unsupervised monocular depth
  estimation with left-right consistency,'' in \emph{Proceedings of the IEEE
  Conference on Computer Vision and Pattern Recognition}, 2017, pp. 270--279.

\bibitem{garg2016unsupervised}
R.~Garg, V.~K. Bg, G.~Carneiro, and I.~Reid, ``Unsupervised cnn for single view
  depth estimation: Geometry to the rescue,'' in \emph{European conference on
  computer vision}.\hskip 1em plus 0.5em minus 0.4em\relax Springer, 2016, pp.
  740--756.

\bibitem{godard2019digging}
C.~Godard, O.~Mac~Aodha, M.~Firman, and G.~J. Brostow, ``Digging into
  self-supervised monocular depth estimation,'' in \emph{Proceedings of the
  IEEE international conference on computer vision}, 2019, pp. 3828--3838.

\bibitem{yin2018geonet}
Z.~Yin and J.~Shi, ``Geonet: Unsupervised learning of dense depth, optical flow
  and camera pose,'' in \emph{Proceedings of the IEEE Conference on Computer
  Vision and Pattern Recognition}, 2018, pp. 1983--1992.

\bibitem{guizilini20203d}
V.~Guizilini, R.~Ambrus, S.~Pillai, A.~Raventos, and A.~Gaidon, ``3d packing
  for self-supervised monocular depth estimation,'' in \emph{Proceedings of the
  IEEE/CVF Conference on Computer Vision and Pattern Recognition}, 2020, pp.
  2485--2494.

\bibitem{johnston2020self}
A.~Johnston and G.~Carneiro, ``Self-supervised monocular trained depth
  estimation using self-attention and discrete disparity volume,'' in
  \emph{Proceedings of the IEEE/CVF Conference on Computer Vision and Pattern
  Recognition}, 2020, pp. 4756--4765.

\bibitem{vijayanarasimhan2017sfm}
S.~Vijayanarasimhan, S.~Ricco, C.~Schmid, R.~Sukthankar, and K.~Fragkiadaki,
  ``Sfm-net: Learning of structure and motion from video,'' \emph{arXiv
  preprint arXiv:1704.07804}, 2017.

\bibitem{zhou2017unsupervised}
T.~Zhou, M.~Brown, N.~Snavely, and D.~G. Lowe, ``Unsupervised learning of depth
  and ego-motion from video,'' in \emph{Proceedings of the IEEE Conference on
  Computer Vision and Pattern Recognition}, 2017, pp. 1851--1858.

\bibitem{luo2018every}
C.~Luo, Z.~Yang, P.~Wang, Y.~Wang, W.~Xu, R.~Nevatia, and A.~Yuille, ``Every
  pixel counts++: Joint learning of geometry and motion with 3d holistic
  understanding,'' \emph{arXiv preprint arXiv:1810.06125}, 2018.

\bibitem{10.1109/TPAMI.2008.132}
A.~Saxena, M.~Sun, and A.~Y. Ng, ``Make3d: Learning 3d scene structure from a
  single still image,'' \emph{IEEE Trans. Pattern Anal. Mach. Intell.},
  vol.~31, no.~5, p. 824–840, May 2009.

\bibitem{Karsch_2014}
K.~Karsch, C.~Liu, and S.~B. Kang, ``Depth transfer: Depth extraction from
  video using non-parametric sampling,'' \emph{IEEE Transactions on Pattern
  Analysis and Machine Intelligence}, vol.~36, no.~11, p. 2144–2158, Nov
  2014.

\bibitem{wu2018sizetodepth}
Y.~Wu, S.~Ying, and L.~Zheng, ``Size-to-depth: A new perspective for single
  image depth estimation,'' 2018.

\bibitem{chen2016singleimage}
W.~Chen, Z.~Fu, D.~Yang, and J.~Deng, ``Single-image depth perception in the
  wild,'' 2016.

\bibitem{Watson_2019_ICCV}
J.~Watson, M.~Firman, G.~J. Brostow, and D.~Turmukhambetov, ``Self-supervised
  monocular depth hints,'' in \emph{Proceedings of the IEEE/CVF International
  Conference on Computer Vision (ICCV)}, October 2019.

\bibitem{Zou_2018}
Y.~Zou, Z.~Luo, and J.-B. Huang, ``Df-net: Unsupervised joint learning of depth
  and flow using cross-task consistency,'' \emph{Lecture Notes in Computer
  Science}, p. 38–55, 2018.

\bibitem{ranjan2019competitive}
A.~Ranjan, V.~Jampani, L.~Balles, K.~Kim, D.~Sun, J.~Wulff, and M.~J. Black,
  ``Competitive collaboration: Joint unsupervised learning of depth, camera
  motion, optical flow and motion segmentation,'' in \emph{Proceedings of the
  IEEE conference on computer vision and pattern recognition}, 2019, pp.
  12\,240--12\,249.

\bibitem{Chen_2019}
Y.~Chen, C.~Schmid, and C.~Sminchisescu, ``Self-supervised learning with
  geometric constraints in monocular video: Connecting flow, depth, and
  camera,'' \emph{2019 IEEE/CVF International Conference on Computer Vision
  (ICCV)}, Oct 2019.

\bibitem{cheng2020s}
B.~Cheng, I.~S. Saggu, R.~Shah, G.~Bansal, and D.~Bharadia, ``$s^3$ net:
  Semantic-aware self-supervised depth estimation with monocular videos and
  synthetic data,'' in \emph{European Conference on Computer Vision}.\hskip 1em
  plus 0.5em minus 0.4em\relax Springer, 2020, pp. 52--69.

\bibitem{shu2020feature}
C.~Shu, K.~Yu, Z.~Duan, and K.~Yang, ``Feature-metric loss for self-supervised
  learning of depth and egomotion,'' in \emph{European Conference on Computer
  Vision}.\hskip 1em plus 0.5em minus 0.4em\relax Springer, 2020, pp. 572--588.

\bibitem{huynh2020guiding}
L.~Huynh, P.~Nguyen-Ha, J.~Matas, E.~Rahtu, and J.~Heikkil{\"a}, ``Guiding
  monocular depth estimation using depth-attention volume,'' in \emph{European
  Conference on Computer Vision}.\hskip 1em plus 0.5em minus 0.4em\relax
  Springer, 2020, pp. 581--597.

\bibitem{xian2020structure}
K.~Xian, J.~Zhang, O.~Wang, L.~Mai, Z.~Lin, and Z.~Cao, ``Structure-guided
  ranking loss for single image depth prediction,'' in \emph{Proceedings of the
  IEEE/CVF Conference on Computer Vision and Pattern Recognition}, 2020, pp.
  611--620.

\bibitem{wang2018non}
X.~Wang, R.~Girshick, A.~Gupta, and K.~He, ``Non-local neural networks,'' in
  \emph{Proceedings of the IEEE conference on computer vision and pattern
  recognition}, 2018, pp. 7794--7803.

\bibitem{zhang2019self}
H.~Zhang, I.~Goodfellow, D.~Metaxas, and A.~Odena, ``Self-attention generative
  adversarial networks,'' in \emph{International conference on machine
  learning}.\hskip 1em plus 0.5em minus 0.4em\relax PMLR, 2019, pp. 7354--7363.

\bibitem{fu2019dual}
J.~Fu, J.~Liu, H.~Tian, Y.~Li, Y.~Bao, Z.~Fang, and H.~Lu, ``Dual attention
  network for scene segmentation,'' in \emph{Proceedings of the IEEE/CVF
  Conference on Computer Vision and Pattern Recognition}, 2019, pp. 3146--3154.

\bibitem{oktay2018attention}
O.~Oktay, J.~Schlemper, L.~L. Folgoc, M.~Lee, M.~Heinrich, K.~Misawa, K.~Mori,
  S.~McDonagh, N.~Y. Hammerla, B.~Kainz, \emph{et~al.}, ``Attention u-net:
  Learning where to look for the pancreas,'' \emph{arXiv preprint
  arXiv:1804.03999}, 2018.

\bibitem{huang2019ccnet}
Z.~Huang, X.~Wang, L.~Huang, C.~Huang, Y.~Wei, and W.~Liu, ``Ccnet: Criss-cross
  attention for semantic segmentation,'' in \emph{Proceedings of the IEEE/CVF
  International Conference on Computer Vision}, 2019, pp. 603--612.

\bibitem{pang2019libra}
J.~Pang, K.~Chen, J.~Shi, H.~Feng, W.~Ouyang, and D.~Lin, ``Libra r-cnn:
  Towards balanced learning for object detection,'' in \emph{Proceedings of the
  IEEE/CVF Conference on Computer Vision and Pattern Recognition}, 2019, pp.
  821--830.

\bibitem{lin2019tsm}
J.~Lin, C.~Gan, and S.~Han, ``Tsm: Temporal shift module for efficient video
  understanding,'' in \emph{Proceedings of the IEEE/CVF International
  Conference on Computer Vision}, 2019, pp. 7083--7093.

\bibitem{santoro2017simple}
A.~Santoro, D.~Raposo, D.~G. Barrett, M.~Malinowski, R.~Pascanu, P.~Battaglia,
  and T.~Lillicrap, ``A simple neural network module for relational
  reasoning,'' \emph{arXiv preprint arXiv:1706.01427}, 2017.

\bibitem{eigen2015predicting}
D.~Eigen and R.~Fergus, ``Predicting depth, surface normals and semantic labels
  with a common multi-scale convolutional architecture,'' in \emph{ICCV}, 2015.

\bibitem{ronneberger2015u}
O.~Ronneberger, P.~Fischer, and T.~Brox, ``U-net: Convolutional networks for
  biomedical image segmentation,'' in \emph{International Conference on Medical
  image computing and computer-assisted intervention}.\hskip 1em plus 0.5em
  minus 0.4em\relax Springer, 2015, pp. 234--241.

\bibitem{wang2018learning}
C.~Wang, J.~Miguel~Buenaposada, R.~Zhu, and S.~Lucey, ``Learning depth from
  monocular videos using direct methods,'' in \emph{Proceedings of the IEEE
  Conference on Computer Vision and Pattern Recognition}, 2018, pp. 2022--2030.

\bibitem{yang2017unsupervised}
Z.~Yang, P.~Wang, W.~Xu, L.~Zhao, and R.~Nevatia, ``Unsupervised learning of
  geometry with edge-aware depth-normal consistency,'' 2017.

\bibitem{mahjourian2018unsupervised}
R.~Mahjourian, M.~Wicke, and A.~Angelova, ``Unsupervised learning of depth and
  ego-motion from monocular video using 3d geometric constraints,'' 2018.

\bibitem{yang2018lego}
Z.~Yang, P.~Wang, Y.~Wang, W.~Xu, and R.~Nevatia, ``Lego: Learning edge with
  geometry all at once by watching videos,'' in \emph{Proceedings of the IEEE
  conference on computer vision and pattern recognition}, 2018, pp. 225--234.

\bibitem{zhan2018unsupervised}
H.~Zhan, R.~Garg, C.~Saroj~Weerasekera, K.~Li, H.~Agarwal, and I.~Reid,
  ``Unsupervised learning of monocular depth estimation and visual odometry
  with deep feature reconstruction,'' in \emph{Proceedings of the IEEE
  Conference on Computer Vision and Pattern Recognition}, 2018, pp. 340--349.

\bibitem{Casser_2019}
V.~Casser, S.~Pirk, R.~Mahjourian, and A.~Angelova, ``Depth prediction without
  the sensors: Leveraging structure for unsupervised learning from monocular
  videos,'' \emph{Proceedings of the AAAI Conference on Artificial
  Intelligence}, vol.~33, p. 8001–8008, Jul 2019.

\bibitem{geiger2013vision}
A.~Geiger, P.~Lenz, C.~Stiller, and R.~Urtasun, ``Vision meets robotics: The
  kitti dataset,'' \emph{The International Journal of Robotics Research},
  vol.~32, no.~11, pp. 1231--1237, 2013.

\bibitem{wang2004image}
Z.~Wang, A.~C. Bovik, H.~R. Sheikh, and E.~P. Simoncelli, ``Image quality
  assessment: from error visibility to structural similarity,'' \emph{IEEE
  transactions on image processing}, vol.~13, no.~4, pp. 600--612, 2004.

\bibitem{pillai2019superdepth}
S.~Pillai, R.~Ambru{\c{s}}, and A.~Gaidon, ``Superdepth: Self-supervised,
  super-resolved monocular depth estimation,'' in \emph{2019 International
  Conference on Robotics and Automation (ICRA)}.\hskip 1em plus 0.5em minus
  0.4em\relax IEEE, 2019, pp. 9250--9256.

\bibitem{karsch2014depth}
K.~Karsch, C.~Liu, and S.~B. Kang, ``Depth transfer: Depth extraction from
  video using non-parametric sampling,'' \emph{PAMI}, 2014.

\bibitem{liu2014discrete}
M.~Liu, M.~Salzmann, and X.~He, ``Discrete-continuous depth estimation from a
  single image,'' in \emph{CVPR}, 2014.

\bibitem{laina2016deeper}
I.~Laina, C.~Rupprecht, V.~Belagiannis, F.~Tombari, and N.~Navab, ``Deeper
  depth prediction with fully convolutional residual networks,'' in \emph{3DV},
  2016.

\bibitem{wang2017learning}
C.~Wang, J.~M. Buenaposada, R.~Zhu, and S.~Lucey, ``Learning depth from
  monocular videos using direct methods,'' in \emph{CVPR}, 2018.

\bibitem{saxena2009make3d}
A.~Saxena, M.~Sun, and A.~Ng, ``Make3d: Learning 3d scene structure from a
  single still image,'' \emph{PAMI}, 2009.

\end{thebibliography}
